\documentclass[letterpaper]{article}
\usepackage{flairs}
\usepackage{times}
\usepackage{helvet}
\usepackage{courier}
\usepackage{makecell}
\usepackage{tikz}
\usepackage{prettyref}
\usepackage{multirow}
\usepackage{url}
\usepackage{listings}
\usepackage{xcolor}
\usepackage{amsmath}
\usepackage{graphicx}

\begin{document}
%

\title{Adversarial Learning to Reason in an Arbitrary Logic}


\author{Stanisław J. Purgał \and Cezary Kaliszyk \\
  \url{{stanislaw.purgal,cezary.kaliszyk}@uibk.ac.at} \\
University of Innsbruck \\
Austria
}

%

\newcommand{\fix}{\marginpar{FIX}}
\newcommand{\new}{\marginpar{NEW}}


\maketitle

\begin{abstract}
Existing approaches to learning to prove theorems focus on particular
logics and datasets. In this work, we propose Monte-Carlo simulations
guided by reinforcement learning that can work in an arbitrarily specified
logic, without any human knowledge or set of problems. Since the algorithm
does not need any training dataset, it is able to learn to work with any
logical foundation, even when there is no body of proofs or even conjectures
available. We practically demonstrate the feasibility of the approach
in multiple logical systems. The approach is stronger than training on randomly generated data but weaker than the approaches
trained on tailored axiom and conjecture sets. It however allows us to apply
machine learning to automated theorem proving for many logics, where no
such attempts have been tried to date, such as intuitionistic logic or linear logic.
\end{abstract}

\section{Introduction}
In the last decade for many logical systems machine learning approaches
have managed to improve on the best human heuristics. This worked well for
example in classical first-order logic guiding the superposition calculus \cite{enigma}, tableaux calculus \cite{mockju-ecai20}, or even in higher-order logic \cite{FarberB16}.
In all these works strategies based on machine learning can significantly improve on the best human-designed
ones.

To train such machine-learned strategies, datasets of problems and baselines on these
problems are required. In particular successful proofs (and in some cases also
unsuccessful proofs~\cite{Kaliszyk2018ReinforcementLO}) are gathered and used to train
a machine-learned version of the prover.

In this work, we consider the same problem but without a fixed logic and without a dataset of
problems given. We apply a policy-guidance algorithm (known for example from AlphaZero \cite{Silver2017MasteringCA}) to proving in an arbitrary logic without a given problem set.
In particular we:
\begin{itemize}
    \item propose a theorem-construction game that allows for learning theorem proving with AlphaZero, without relying on training data;
    \item propose the first dataset for learning for multiple logics, together with learning baselines
    for this dataset.
    \item propose an adjusted Monte-Carlo Tree Search that is able to take into account certain (sure) information, when a player makes multiple moves (explained in ``Certain Value Propagation'' section); 
    \item evaluate the trained prover on the dataset
    showing that it improves proof capability in various
    considered
    logics; and
    \item as many other problems and games can be directly encoded as logical problems we show that the proposed universal learning for logic also works on some encoded games, such as Sokoban.
\end{itemize}


\section{Related Work} \label{s:related}

\cite{Silver2017MasteringCA} have shown that Monte-Carlo tree search combined with reinforcement learning applied to policy and value functions can generalize to multiple logical games (Go, Chess, Shogi). 

The work of \cite{firoiu2021training} focuses on classical first-order logic without equality and the resolution calculus (so only one of the many logics we consider) and (like us) does not use the complete TPTP problems, but only the axiom sets to learn. There is however a significant overlap between the axioms and conjectures in other problems. The resulting prover does learn to prove theorems but is significantly weaker than E-prover on the TPTP problems. The idea to use only the axioms has already been considered by \cite{Bansal2019LearningTR}. Even if no original conjectures are exposed to the prover, the dataset used for training is quite large, in comparison with ours, where no formulas are given at all.

As already discussed in the introduction, there are many approaches to applying MCTS with policy and guidance
learning in various fixed calculi and on fixed datasets \cite{Kaliszyk2018ReinforcementLO,Rawson2019ANP}. The results are better than those we are able to get here, but no new logics or problems are tried and generalization and transfer have been very limited so far.
The AlphaZero algorithm has also been applied in theorem proving to the synthesis of formulas \cite{brown2019selflearned} and functions \cite{gauthier2019deep}. 

Kaiser et al. demonstrated that theorem proving can be used to solve many games, such as the ones in the General Game Playing competition \cite{DBLP:conf/aaai/KaiserS11}. With the current work, we show that learning for logic can be also applied to these games. 

\section{Preliminaries}\label{s:prelim}

\subsection{AlphaZero}
The core of the AlphaZero algorithm \cite{Silver2017MasteringCA} is learning from self-play. It trains a neural network to evaluate the states of a game to estimate the final outcome of the game as well as a policy maximizing the expected outcome.
Using the neural network in the current stage of learning a lot of playouts are generated, then this data is used as training data for further improvement.

For training value estimation, the algorithm uses the actual outcomes of the games. To train policy estimation, a Monte-Carlo Tree Search (MCTS) is used to compute a better policy, then the network is trained to return this better policy.

\subsection{Monte-Carlo Tree Search}
\label{sec:mcts}
For the purpose of training the policy evaluating network we need to provide it with a somewhat better policy. This is done by exploring a tree of possible moves. It is a guided exploration, biased towards the moves pointed to by the policy and to where the value estimations are higher.

A tree is constructed with every node representing a state of the game. Each of those states is evaluated using the neural network. When deciding where to add a new node (thus exploring a branch of the game further) the MCTS algorithm takes into account both the value and the policy estimations from the neural network (biased toward following the policy and higher values), as well as how well a branch was already explored (biased to explore yet unexplored branches more).

After adding a set amount of nodes to the tree, the new better policy is defined to be proportional to the number of nodes explored below each of the immediate children of the root node (representing the state for which we are computing a better policy).


\section{Approach}\label{s:approach}

\subsection{The theorem-construction game}
\label{sec:game}
We propose a two-player game such that a system trained to play the game well could be used to effectively prove the theorems of a given logic system.
The first player (referred to as \emph{adversary}) constructs a provable statement, while the other player (referred to as \emph{prover}) tries to prove it. The goal of the adversary is to construct such a theorem, that the prover will fail to prove it. However, because of the available game moves (construction steps), the statements are always provable.

\begin{figure}[htb]
    \centering
    \begin{tikzpicture}[scale=0.6]
    \draw[thick,->] (2,2) -> (2,1);
    \draw[fill=white] (-1.5,0) rectangle (5.5,1) node[text=black,pos=.5] {\small \emph{Adversary} constructs a theorem};
    \draw[thick,->] (2,0) -> (2,-1);
    \draw[fill=black] (-1.5,-2) rectangle (5.5,-1) node[text=white,pos=.5] {\small \emph{Prover} proves the theorem};
    

    \node[text width=2cm] at (-.8, 0.9) {\small fail};
    \node[text width=2cm] at (4, -0.3) {\small succeed};

    \node[text width=2cm] at (4, -2.3) {\small succeed};
    \node[text width=2cm] at (7.4, -1.1) {\small fail};
    

    \draw[fill=gray] (5,-3.5) rectangle (9,-2.5) node[text=black,pos=.5] {\small \emph{Adversary} wins};
    \draw[fill=gray] (-4,-3.5) rectangle (-1,-2.5) node[text=white,pos=.5] {\small \emph{Prover} wins};
    
    \draw[thick,->] (2,-2) |- (-1,-3);
    \draw[thick,->,red] (-1.5,0.5) -| (-2.5,-2.5);
    \draw[thick,->,red] (5.5,-1.5) -| (7.5,-2.5);
    \end{tikzpicture}
    \caption{High-level overview of the theorem-construction game.}
    \label{fig:game_overview}
\end{figure}
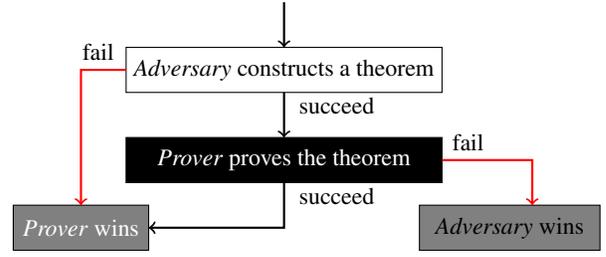

We represent the game objects using Prolog-like \emph{terms}, where a term can be either a \emph{variable} or a pair of an atom and a list of subterms. In the examples, we use the convention of marking variables with capital letters, and denoting compound terms as an atom name followed by a list of subterms in brackets (skipped when the list is empty). For example
\(
\texttt{tee}(A, \texttt{implies}(B, \texttt{false}))
\), which can also be expressed with operators like 
\begin{math}
(A \vdash (B \to \bot))
\end{math}.
 
The construction game is defined for a given set of \textit{inference rules}. An inference rule is a pair of a term and a list of terms, that can share variables. For example
\(
\texttt{tee}(A, \texttt{and}(B, C)) \leftarrow \texttt{tee}(A, B), \texttt{tee}(A, C)
\), equivalent to
\(
(A \vdash (B \land C)) \leftarrow (A \vdash B), (A \vdash C)
\).

For the prover, a game state consists of a list of terms that need to be proven (together with the information that the prover is making the move). During their move, a player can choose one of the given inference rules (the \emph{action space} is the set of \emph{inference rules}), and apply it to the first term of the list. The left side of the rule is then unified with that term. If the unification fails, the player making the move loses. If it succeeds, the term is removed from the list, and the right side of the rule (after unification) is added.

The \textit{adversary} (the player constructing a theorem) makes moves in much the same way, except instead of starting with a theorem to be proven, it starts with a single variable. 
Applying inference rules to prove this variable will unify it with some term. If the proof is successfully completed, this variable we started with will be unified with a provable theorem. We keep track of this variable in the game state. When the adversary finishes its proof, we pass the constructed theorem to the prover, after replacing all remaining variables with fresh constants.

The second player  tries to prove the theorem, winning when the list is empty.
To better illustrate the working of our theorem-construction game we present the rules of a concrete game in Figure \ref{fig:example_rules} and an example playout in Figure \ref{fig:example_game}.

\begin{figure}[tbh]
             \begin{align}
                 A,B \vdash A &\leftarrow \\
                 A \vdash (B \to C) &\leftarrow (B,A \vdash C) \\
                 A \vdash (B \land C) &\leftarrow (A \vdash B), (A \vdash C) \\
                 A \vdash B &\leftarrow (A \vdash (B \land C)) \\
                 A \vdash B &\leftarrow (A \vdash (C \land B)) \\
                 A \vdash B &\leftarrow (A \vdash \bot)
             \end{align}
    \caption{\label{fig:example_rules}A subset of propositional logic inference rules used in the example in figure \ref{fig:example_game}}
    
\end{figure}

\begin{figure*}
\centering
        \begin{tabular}{r|c|l}
            rule & terms to be proven & constructed theorem  \\
            \hline
            & $X$ & $X$  \\
            2 & $A,B \vdash C$ & $B \vdash A \to C$ \\
            3 & $(A,B \vdash D), (A,B \vdash E)$ & $B \vdash A \to (D \land E)$ \\
            4 & $(A,B \vdash (D \land F)), (A,B \vdash E)$ & $B \vdash A \to (D \land E)$ \\
            1 & $(D \land F),B \vdash E$ & $B \vdash (D \land F) \to (D \land E)$ \\
            6 & $(D \land F),B \vdash \bot$ & $B \vdash (D \land F) \to (D \land E)$ \\
            5 & $(D \land F),B \vdash (G \land \bot)$ & $B \vdash (D \land F) \to (D \land E)$ \\
            1 & & $B \vdash (D \land \bot) \to (D \land E)$  \\
             \hline
            & $\texttt{b} \vdash (\texttt{d} \land \bot) \to (\texttt{d} \land \texttt{e})$ & \\ 
            2 & $(\texttt{d} \land \bot),\texttt{b} \vdash (\texttt{d} \land \texttt{e})$ &  \\
            6 & $(\texttt{d} \land \bot),\texttt{b} \vdash \bot$ &  \\
            5 & $(\texttt{d} \land \bot),\texttt{b} \vdash (A \land \bot)$ &  \\
            1 & \multicolumn{2}{|c}{\text{Prover won}} \\
        \end{tabular}
    \caption{\label{fig:example_game}
    An example of a playout of the theorem-construction game with the inference rules shown in figure \ref{fig:example_rules}}
    
\end{figure*}

\subsection{Certain Value Propagation}
\label{sec:cvp}
The AlphaZero \cite{Silver2017MasteringCA} algorithm utilizes a neural network to estimate state values (a number in range $(-1, 1)$, we will call it $v_\theta$) and policies (a vector with as many dimensions as the size of the action space). Then a Monte Carlo Tree Search (MCTS) \cite{kocsis2006bandit} is used to compute better estimates of value and policy. This is done by exploring the tree of possible playouts, with a bias toward where value and policy lead to.

During this exploration, MCTS maintains a better value estimation of every state (we will refer to it as $v$), which is defined to be the average of $v_\theta$ of all explored descendants. We will use an equivalent definition, as the weighted average of immediate children, with weights being the number of visits of a given node (the difference will become important).
\begin{align*}
    v(n) &= \frac{v_\theta(n) + \sum_{d < n} v_\theta(d)}{| d : d < n |+1} = \\
    &= \frac{v_\theta(n) + \sum_{c \in \text{children}(n)} v(c) * (|{d : d < c}| + 1)}{|{d : d < n}|+1}
\end{align*}

In our version of MCTS, for every node we keep track of a lower and upper bound for possible node values. For non-final nodes, these are simply ($-1$, $1$) (as this is the range of possible outcomes), but for the final nodes, the bounds are both equal to the final reward. These bounds are propagated up the tree in a natural way (taking into account state ownership). Then, for every node we compute a new $v_c$ value, which is simply $v$ adjusted to fall within lower-upper bounds -- so eg. if the lower bound is higher than $v$, then $v_c$ will be equal to the lower bound. Then, when computing $v$ for the nodes above we use this new $v_c$ value rather than the old $v$.
\begin{align*}
    v_c(n) &= \max(\text{lower}(n), \min(\text{upper}(n), v(n))) \\
    v (n) &= \frac{v_\theta(n) + \sum_{c \in \text{children}(n)} v_c(c) * (|{d : d < c}| + 1)}{|{d : d < n}|+1}
\end{align*}.

\usetikzlibrary{automata,positioning}

\begin{figure}[htb]
    \centering
    \begin{tikzpicture}[scale=0.8,every node/.style={font=\small}]
    \draw[thick,-] (0,0) -> (0,1.2);
    \draw[thick,-] (0,0) -> (-1,-2);
    \draw[thick,-] (0,0) -> (1,-2);
    \draw[thick,-] (1,-2) -> (0,-4);
    \draw[thick,-] (1,-2) -> (2,-4);
    \node [draw,fill=cyan, align=center,shape=rectangle,minimum width=1.4cm,minimum height=1.5cm] at (0,0) {$v= - 0.46$ \\ $v_\theta = -0.1$};
    \node [draw,fill=cyan,align=center,shape=rectangle,minimum width=1.4cm,minimum height=1.5cm] at (-1,-2) {$v= - 0.6$ \\ $v_\theta = - 0.6$};
    \node [draw,fill=pink,align=center,shape=rectangle,minimum width=1.4cm,minimum height=1.5cm] at (1,-2) {$v=0.5$ \\ $v_\theta = 0.3$};
    \node [draw,fill=pink,align=center,shape=rectangle,minimum width=1.4cm,minimum height=1.5cm] at (0,-4) {$v=0.2$ \\ $v_\theta = 0.2$};
    \node [draw,fill=pink,align=center,shape=rectangle,minimum width=1.4cm,minimum height=1.5cm] at (2,-4) {$v=1$ \\ (final)};
    \end{tikzpicture}
    \begin{tikzpicture}[scale=0.8,every node/.style={font=\small}]
    \draw[thick,-] (0,0) -> (0,1.2);
    \draw[thick,-] (0,0) -> (-1,-2);
    \draw[thick,-] (0,0) -> (1,-2);
    \draw[thick,-] (1,-2) -> (0,-4);
    \draw[thick,-] (1,-2) -> (2,-4);
    \node [draw,fill=cyan,align=center,shape=rectangle,minimum width=1.4cm,minimum height=1.5cm] at (0,0) {$v=- 0.76$ \\ $v_\theta = 0.1$ \\ $(-1, 1)$ \\ $v_c = -0.7$};
    \node [draw,fill=cyan,align=center,shape=rectangle,minimum width=1.4cm,minimum height=1.5cm] at (-1,-2) {$v= - 0.6$ \\ $v_\theta = - 0.6$ \\ $(-1, 1)$ \\ $v_c = -0.6$};
    \node [draw,fill=pink,align=center,shape=rectangle,minimum width=1.4cm,minimum height=1.5cm] at (1,-2) {$v=0.5$ \\ $v_\theta = 0.3$ \\ $(1, 1)$ \\ $v_c = 1$};
    \node [draw,fill=pink,align=center,shape=rectangle,minimum width=1.4cm,minimum height=1.5cm] at (0,-4) {$v=0.2$ \\ $v_\theta = 0.2$ \\ $(-1, 1)$ \\ $v_c = 0.2$};
    \node [draw,fill=pink,align=center,shape=rectangle,minimum width=1.5cm,minimum height=1.5cm] at (2,-4) {$v=1$ \\ (final) \\ $ (1,1) $ \\ $v_c = 1$ };
    \end{tikzpicture}
    \caption{Examples of value estimation in MCTS without Certain Value Propagation (left) and with (right). State ownership (which player is making a move) is marked with color.}
    \label{fig:mcts_cvp}
\end{figure}
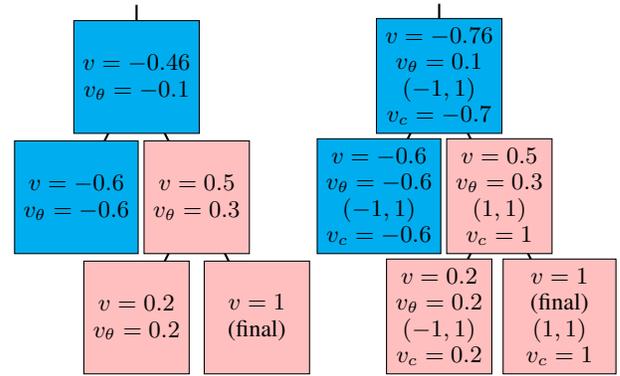

Additionally, whenever the value estimation of a state is determined to be $-1$ (the lowest possible outcome), this state will be avoided. This avoidance is applied both to MCTS exploration and the choice of an action to take during playouts.

The impact of certain value propagation on the final performance of the prover is shown in Fig. \ref{fig:mods}.

\subsection{Auxiliary replays}
\label{s:aux_replays}
To facilitate the prover learning to prove theorems constructed by the adversary we add additional \emph{auxiliary replays}. These come from the  games won by the adversary when the prover fails to prove a constructed theorem. Because of the way the theorem was constructed we know how to prove it -- we just need to apply the same moves that the adversary used to construct it. Using this fact, we create a replay that shows how the theorem could be proven. In this replay, the policy is not computed using MCTS, but rather is just a one-hot vector pointing to the move that the adversary made when constructing the theorem.

With these auxiliary replays, our algorithm can be considered to train on artificially constructed theorems, that at first come from simply randomly applying inference rules, but later on uses neural guidance to find theorems that the prover cannot yet prove. However, as mentioned earlier, we only do this for theorems that the prover failed to prove. 

The impact of including auxiliary replays on the final performance of the prover is shown in figure \ref{fig:mods}.

\subsection{Balancing training data}
Since the theorem proving game (explain in section ``The theorem-construction game'') is asymmetrical, simply using all replay data for training would result in an unbalanced dataset. On top of this, we use auxiliary replays (explained in section ``Auxiliary replays''), further disturbing the training data.

To deal with this imbalance we apply training data balancing.
Replays are split into parts according to which player won, and a third set of auxiliary replays. All training batches contain the same number of examples from each part.

However, this means disturbing the way the Mean Square Error loss works for value estimation. Consider a value estimation of the starting state. Normally, optimal loss for it would be achieved if the estimate was the average outcome of the game, but with balancing the optimal loss will be achieved by estimating the value to be $0$ (mean between losing and winning). This problem affects every state that occurs multiple times in the training dataset.

To counteract this problem, the value loss is weighted in proportion to the size of the part of the data, from which the point originates. So if a player $A$ won in proportion $4:1$, the training batches would include games won by this player in proportion $1:1$, but $\frac{4}{5}$ of the loss (and therefore gradients) would be determined by data from games won by the player $A$. For auxiliary replays, this weight is set to $0$ and only policy is learned from them.

\begin{figure}[htb]
    \centering
    \includegraphics[width=\columnwidth]{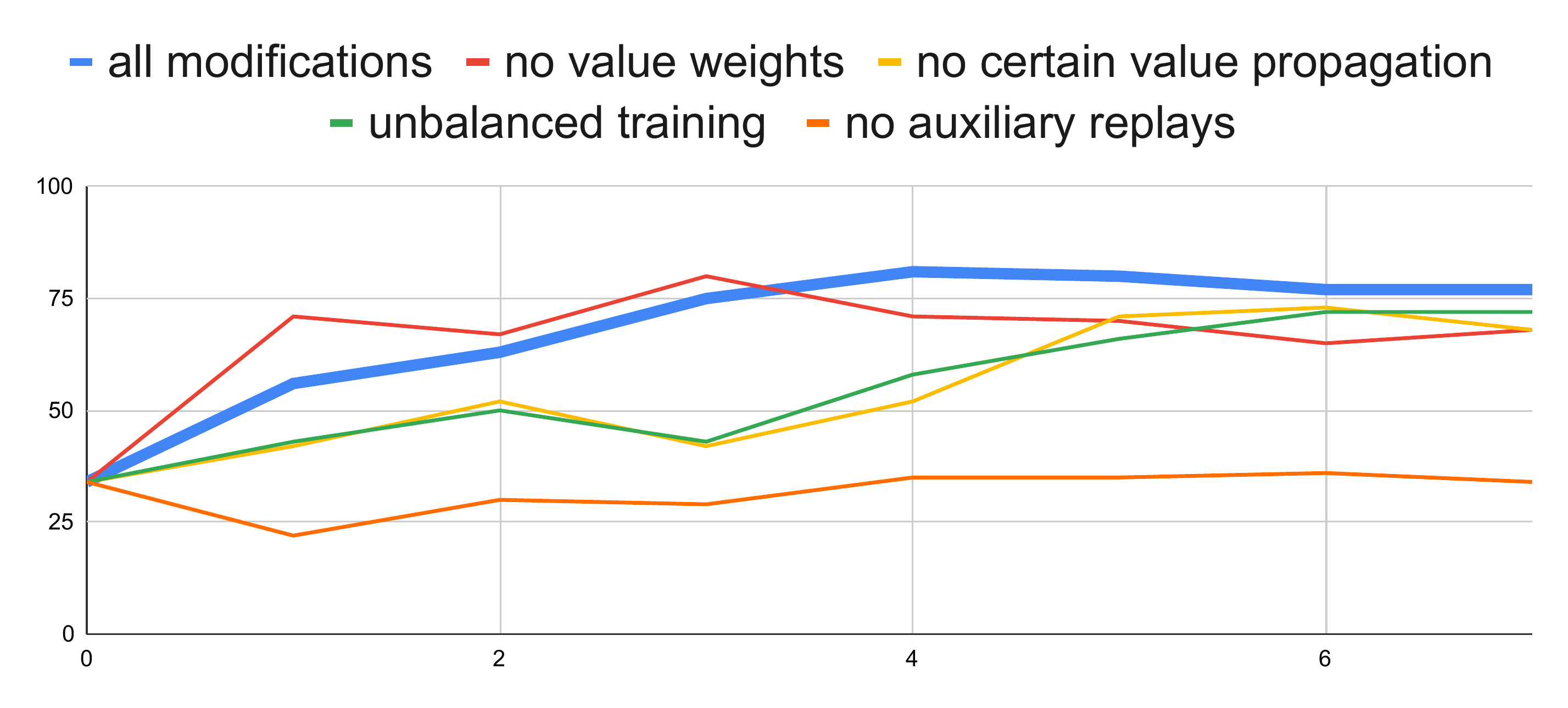}
    \caption{The impact of our modifications on our algorithm. Solved first-order classical tableaux test problems over time (episodes) with 5k games per episode.}
    \label{fig:mods}
\end{figure}

The impact of balancing training data and weighing the value loss on the final performance of the prover is shown in figure \ref{fig:mods}.

\subsection{Applicability}
As mentioned in section ``The theorem-construction game'', our system works with a logic system defined by a set of inference rules. This set of rules can be thought of as a Prolog program, and since Prolog is Turing-complete our method can (at least theoretically) be used to learn to reason in any formally defined (and decidable) context. As an example of wide applicability, we train our system to solve Sokoban puzzles. This can be done by defining rules of the game as inference rules of a pseudo-``logic system''.

This of course does not mean that the system will always work well. For example, a saturation prover requires all terms used in the proof to be fully determined from the start. This negates the advantage of the adversary player, who normally can still modify the constructed theorem late into its proof. Because of this, the probability of constructing a theorem that an untrained prover cannot prove becomes really low -- so low that potentially no such theorems will be generated for the initial training set. In such a situation, nothing can be learned from such data and the system is stuck. This problem could potentially be overcome by simply generating enough playouts, but in our experiments with using a saturation prover, the system got stuck after the first step, with the prover winning all games. This was the case in our experiments using a saturation proving method, with $10^4$ games generated per episode (possibly more games could help).

\subsection{Failure states}
Another consequence of the game being asymmetrical is the possibility of the training getting stuck when one player starts winning every time.
The mechanism of auxiliary replays mentioned above counteracts this to some extent, allowing the prover to still learn even if the adversary is always successfully constructing a hard enough theorem.
If the prover was winning every game, however, we would need to rely on exploration for the adversary to find something hard to prove.
This situation is however virtually impossible, because of the exploration noise used during playouts. This should lead to adversary towards theorems where the prover is uncertain and sometimes loses due to exploration noise, and then to theorems where the prover fails.

There is however another failure state which if reached would be entirely stable. It is possible because the construction of theorems is inherently easier than proving. Consider a theorem $\exists_x \textsc{hash}(x) = y$. It is easy to prove such a theorem if one can choose what $y$ is going to be. If $y$ is already decided, proving such a theorem becomes extremely difficult. So difficult in fact, that we cannot hope that a neural network would be able to learn to do this.

If the adversary found such a space of \emph{Uninteresting Hard Theorems}, it would never learn to do anything else. After all, it is a winning strategy for this game. The prover, even using the auxiliary replays would never learn to do anything useful in this situation, and would gradually forget all the useful knowledge learned previously.

This does not seem to happen in any of our experiments. In some of our considered logic systems, it is not even clear that such an Uninteresting Hard Theorem space exists.

\subsection{Neural architecture}
\label{sec:neural_arch}
For evaluating state value and policy we use a Graph Neural Network similar to the one described in \cite{expgnn}. It is essentially a Graph Attention Network \cite{graph_attention} using dot-product attention from the Transformer model \cite{vaswani2017attention} with different attention masks for different attention heads.
One Graph Neural Network is used to create a single vector representation of the graph, which is then fed to the final layers to estimate policy and value.

\includegraphics[width=1.0\columnwidth]{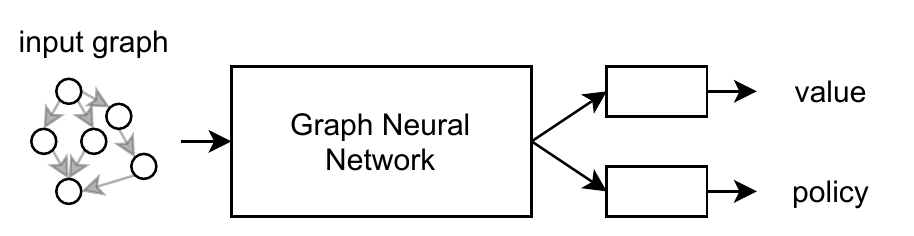}

\begin{figure}[htb]
    \centering
    \includegraphics[width=0.7\columnwidth]{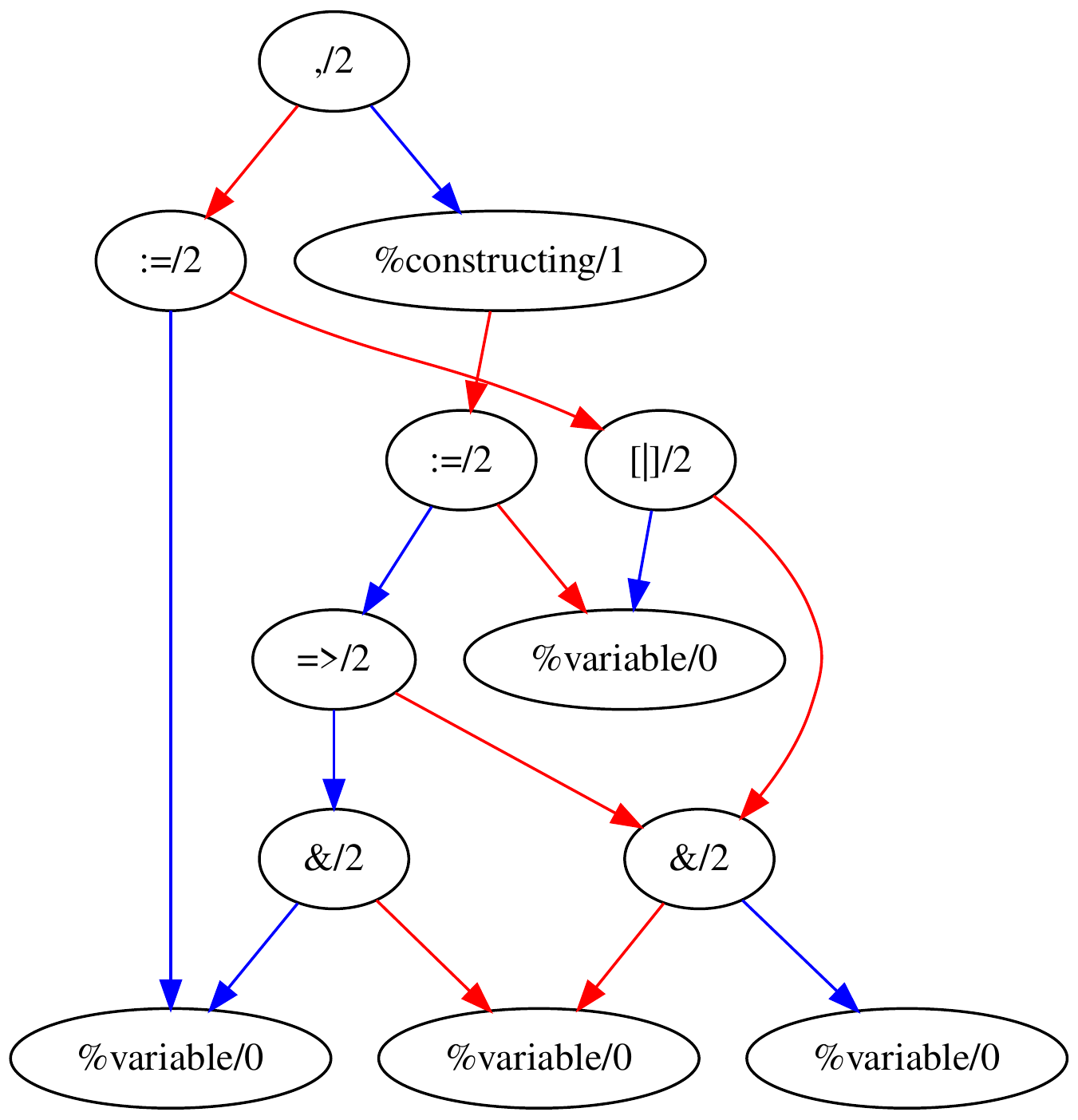}
    \caption{An example graph representing the game state from the game in figure \ref{fig:example_game} after initial 4 moves. }
    \label{fig:example_graph}
\end{figure}

Game states are represented as syntactic graphs. One graph contains all terms that need to be proven, together with information about which player the state belongs to, and (for the adversary player) the state of the constructed theorem. An example of such a graph is shown in figure \ref{fig:example_graph}.

A single Graph Neural Network is used to evaluate the states for both players, the prover and the adversary.

\section{Evaluation}\label{s:eval}

To test the impact of our method of adversarial training we compare an algorithm trained using our theorem-construction game with a prover trained using uniformly generated random data (an approach somewhat similar to \cite{firoiu2021training}).

The methods are tested on a dataset not seen by either approach. This test dataset is human-generated (see section ``Considered logics'' for details on each test dataset) and is often very far outside the training distribution.

\subsection{Baseline}
We generate baseline training data by applying inference rules randomly. This is essentially the adversary from our game doing random moves. Because this does not require evaluating states with a neural network, generating such data is much cheaper, so we generate more playouts -- $10^6$ (we note that not all playouts result in a constructed theorem).

We use all data generated this way to train a network to estimate policy and value. The policy is a one-hot vector pointing to what the adversary did to construct a theorem, and the value is $0.99^n$ with $n$ being the number of moves left to do.

\begin{table}[htb]
\begin{tabular}{r|c|c|c}
\multicolumn{1}{c|}{logic}              & 
\begin{tabular}[c]{@{}c@{}}baseline\end{tabular} &
\begin{tabular}[c]{@{}c@{}}best \\ during \\ game \\ training\end{tabular} &
\begin{tabular}[c]{@{}c@{}}total \\ solved \\ during \\ training\end{tabular}
\\ \hline
int. prop. sequent & 12  & 12 & \bf 13
\\ \hline
classical FO sequent & \bf 42 & 39 & 40                                              \\ \cline{2-4} 
classical FO tableaux  & 73 & 79 & \bf 83                                            \\ \cline{2-4} 
classical FO Hilbert           & \bf 38  & 37 & \bf 38                                   \\ \hline
modal K prop. sequent        & 2 & 5 & \bf 7
\\ \cline{2-4} 
modal T prop. sequent        & 6 & 12 & \bf 13                                         \\ \cline{2-4} 
modal S4 prop. sequent       & 2 & 6 & \bf 8                                         \\ \cline{2-4} 
modal S5 prop. sequent       & 4 & 24 & \bf 24                                        \\ \hline
linear prop. sequent         & 37 & 34 & \bf 39
\\ \hline
sokoban solving         & 4  & 10 & \bf 12                                     
\end{tabular}
\label{tab:all}
\caption{Results of training in all tested logics -- \emph{int.} and \emph{prop.} stand for intuitionistic and propositional }
\end{table}

\subsection{Setup}
We implemented the proposed theorem-construction game engine
SWI-Prolog \cite{wielemaker2010swiprolog}
using  PyTorch \cite{paszke2019pytorch}
for the proposed adversarial neural architecture.

We train our system in episodes, first generating $10^4$ playouts, then training the neural network using these playouts as training data. This step is repeated multiple times, and after every one, we evaluate the system using the test dataset.

\subsection{Experiments}
To test how much the prover has learned, we play the game in a similar way to when training, except skipping the construction phase, and instead using a theorem from the test set. During such testing we forgo forcing exploration -- we do not add exploration noise in Monte-Carlo Tree Search and use the most probable action instead of choosing randomly. Also, when a final state is found during MCTS exploration, we just follow a path to it.

For termination during testing we use a limit on explored states -- nodes added to the MCTS tree. Because a part of the tree can be reused for the next state (the part below the node that was chosen) this does not imply any strict turn limit. 

\subsection{Considered logics}

\paragraph{Intuitionistic}


We train our prover on sequent calculus in propositional intuitionistic logic \cite{HuthRyan}. For test theorems, we use a part of the ILTP library \cite{raths2005iltp}.

\paragraph{Classical}

We run three experiments with classical first-order logic, trying out three different proof systems. One is sequential calculus, the same as used with intuitionistic logic, another is the Tableaux connection prover \cite{Hahnle01}, and lastly the rather unwieldy Hilbert system. For the test set, we use a small subset of the Mizar40 dataset~\cite{ckju-miz40-jar15} of formulas that do not use equality. 

\paragraph{Linear}
We also train in linear logic \cite{girard1987linear}, only in the propositional setting. For the evaluation we use the LLTP \cite{lltp} library, most of which is taken from ILLTP \cite{olarte2019illtp}. We also use a few hand-written examples.

\paragraph{Modal}
In another experiment we train the prover to work with modal propositional logic \cite{BBW2007}, in four variants: K, T, S4, S5. Each of those extends
the definition of the logic by an additional rule.

For evaluation we use the propositional part of the QMTLP library \cite{raths2011qmltp}. The set of test theorems is expanded for each consecutive added rule.

\begin{figure}
    \centering
    \includegraphics[width=0.23\columnwidth]{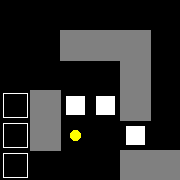}
    \includegraphics[width=0.23\columnwidth]{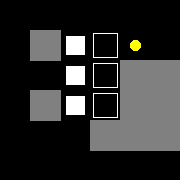}
    \includegraphics[width=0.23\columnwidth]{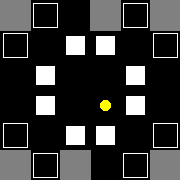}
    \includegraphics[width=0.23\columnwidth]{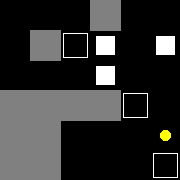}
    \caption{A few examples of Sokoban problems}
    \label{fig:sokoban_examples}
\end{figure}

\paragraph{Sokoban}
Sokoban is a classic puzzle game, where the goal is to push boxes into their target positions. The puzzle is PSPACE-complete \cite{culberson1997sokoban}.
We only generate puzzles of size 6 by 6.
For testing we use a dataset available online\footnote{\url{https://sourceforge.net/projects/sokoban-solver-statistics/}} (only the problems that can fit into a 6 by 6 grid). A few examples of such problems are shown in figure \ref{fig:sokoban_examples}.

\section{Results and Discussion}
The results of the evaluation are presented in table 1. We compare the baseline against the last model trained in the adversarial setup, and additionally list the number of unique problems solved in all training epochs. For all modal logics as well as for classical first-order Tableaux the proposed adversarial theorem-construction game leads to many more solved problems.  For a number of other logical calculi, the adversarial version is slower but leads to finding solutions different than those trained in the supervised setting, therefore leading to a large number of total solutions found. This is the case for intuitionistic sequent calculus and linear logic. Among the tried calculi, only for the classical sequent-calculus and Hilbert-calculus there is no advantage - this is likely due to the fact that the calculus is much closer to the syntax and the learned baseline can generalize enough. Finally, for the encoded Sokoban games 
the results are particularly good, with many games 
solved only in the adversarial-logical setting.
We believe, that the adversary learns to construct more and more complex Sokoban-encoded proof games, while the player learns to solve them, in a way similar to curriculum learning.

\subsection{Forwards vs. backwards conjecturing}
During our experiments we briefly considered implementing a different method of constructing theorems, namely forward constructing: starting from assumptions, working towards the theorem. This method is used in \cite{firoiu2021training} to generate synthetic data (though without any training for the generator).

The problem with forward constructing (and the reason we decided not to use it) can be illustrated using the following inference rule (disjunction elimination):
\begin{align*}
(T \vdash C) \leftarrow&\ (T \vdash A \lor B), (T \vdash A \to C), (T \vdash B \to C)
\end{align*}

For the adversary to ever successfully apply such a rule, it would first have to construct three statements specifically fitting the rule. In the initial phase of random exploration that would be extremely unlikely. Moreover, the policy predictor will quickly learn that applying this rule always ends in failure, thus making its use even more unlikely after some training. In backwards construction the rule can easily be applied, and simply results in three new statements that consequently need to be proven.

This problem essentially does not exists in the case of a saturation prover, which uses a single inference rule that can be applied anywhere.

\subsection{Comparison with existing methods}
Saturation provers are the state-of-the-art for first-order theorem proving are. These are designed narrow down the search space compared to possibly applying any inference rule at any point. Moreover, their solutions often involve millions of inference steps, while in our case the limit of moves is in the order of $10^2$. As such many problems from the test dataset may not even technically be solvable by our system.

For these reasons, our system performs a lot worse than these in their respective domains. It can however be applied to any formally defined domain and is (as far as the authors know) the only proposed theorem-proving system that may continuously learn and improve without any dataset.

\section{Conclusions}
We presented an algorithm for learning to reason in an arbitrary logic. The system, given only a formal definition of a logic, learns to construct increasingly harder problems in the logic and learns to prove them. We show that the system does learn to perform better than a baseline system trained using uniformly generated logical problems. The performance is of course weaker than that of domain-specific Automated Theorem Provers and provers trained on tailored datasets. 
 We are, however, able to construct automatically the first efficient learned automated theorem provers for some logics where none existed before, including various modal logics.

Future work includes encoding more intricate theorem proving calculi, in order to compare them with the
more tailored machine-learned systems.
Furthermore, for most of the considered logics, the performance on the test sets has stagnated after a few episodes. It remains an open question if trying a compute power comparable with AlphaZero \cite{Silver2017MasteringCA} would produce significantly better results.

\paragraph{Acknowledgements}

This work has been supported by the ERC starting grant no. 714034 SMART.


\end{document}